%% file: acl_latex.tex
\pdfoutput=1
\documentclass[11pt]{article}
\PassOptionsToPackage{table}{xcolor}
\usepackage[preprint]{acl}
\usepackage{times}
\usepackage{latexsym}
\usepackage{booktabs}
\usepackage{graphicx}
\usepackage{amsmath}
\usepackage{pifont}
\usepackage{xspace}
\usepackage{amssymb}
\usepackage{enumitem}
\usepackage{tabularx}
\usepackage{multirow}

\usepackage[T1]{fontenc}

\usepackage[utf8]{inputenc}
\usepackage{microtype}

\DeclareUnicodeCharacter{211D}{\ensuremath{\mathbb{R}}}
\DeclareUnicodeCharacter{2124}{\ensuremath{\mathbb{Z}}}
\DeclareUnicodeCharacter{21A6}{\ensuremath{\mapsto}}

\DeclareUnicodeCharacter{2264}{\ensuremath{\leq}}
\DeclareUnicodeCharacter{2265}{\ensuremath{\geq}}
\DeclareUnicodeCharacter{2192}{\ensuremath{\rightarrow}}
\DeclareUnicodeCharacter{2194}{\ensuremath{\leftrightarrow}}
\DeclareUnicodeCharacter{2227}{\ensuremath{\wedge}}
\DeclareUnicodeCharacter{22A2}{\ensuremath{\vdash}}
\DeclareUnicodeCharacter{2080}{\ensuremath{_0}}
\DeclareUnicodeCharacter{2081}{\ensuremath{_1}}
\DeclareUnicodeCharacter{2082}{\ensuremath{_2}}
\DeclareUnicodeCharacter{2083}{\ensuremath{_3}}
\DeclareUnicodeCharacter{2084}{\ensuremath{_4}}
\DeclareUnicodeCharacter{2085}{\ensuremath{_5}}
\DeclareUnicodeCharacter{2086}{\ensuremath{_6}}
\DeclareUnicodeCharacter{2087}{\ensuremath{_7}}
\DeclareUnicodeCharacter{2088}{\ensuremath{_8}}
\DeclareUnicodeCharacter{2089}{\ensuremath{_9}}
\DeclareUnicodeCharacter{2200}{\ensuremath{\forall}}
\DeclareUnicodeCharacter{2203}{\ensuremath{\exists}}
\DeclareUnicodeCharacter{2228}{\ensuremath{\lor}}
\DeclareUnicodeCharacter{2260}{\ensuremath{\neq}}
\DeclareUnicodeCharacter{230B}{\ensuremath{\rfloor}} 
\DeclareUnicodeCharacter{2223}{\ensuremath{\mid}}    
\DeclareUnicodeCharacter{230A}{\ensuremath{\lfloor}} 
\DeclareUnicodeCharacter{2090}{\ensuremath{_a}}  
\DeclareUnicodeCharacter{2091}{\ensuremath{_e}}  
\DeclareUnicodeCharacter{2092}{\ensuremath{_o}}  
\DeclareUnicodeCharacter{2093}{\ensuremath{_x}}  
\DeclareUnicodeCharacter{2211}{\ensuremath{\sum}}        
\DeclareUnicodeCharacter{2115}{\ensuremath{\mathbb{N}}}  

\DeclareUnicodeCharacter{2208}{\ensuremath{\in}} 

\newcommand{\benchmark}{FinForesight}
\definecolor{tableHeader}{HTML}{EEF2F7}
\newcommand{\bestscore}[1]{\textbf{#1}}
\title{\benchmark: A Diagnostic Backtesting Benchmark for Deep Research Systems in Financial Forecasting}

\author{%
    \textbf{
    Fengyuan Liu$^{1,5}$\thanks{Equal contribution.} \hspace{0.5em}
    Mukai Li$^{1,5}$\footnotemark[1]\thanks{Corresponding author.} \hspace{0.5em}
    Jieshun You$^{2}$\footnotemark[1] \hspace{0.5em}
    Yilin Zheng$^{1}$\footnotemark[1] \hspace{0.5em}
    Chenjia Min$^{3}$ \hspace{0.5em}
    }\\
    \textbf{
    Yizhou Zhang$^{4,6}$ \hspace{0.5em}
    Harper Chen$^{4,6}$ \hspace{0.5em}
    Qi Liu$^{1,5}$\footnotemark[2]
    }\\[3mm]
    $^{1}$The University of Hong Kong \hspace{0.5em}
    $^{2}$Hong Kong Polytechnic University \\
    $^{3}$University of Cambridge \hspace{0.5em}
    $^{4}$Lanzhou University \\
    $^{5}$Grace Investment Machine \hspace{0.5em}
    $^{6}$Haitun Information Technology \\
    \small\texttt{\{fengyuanhku, limukai.nlp\}@connect.hku.hk, liuqi@cs.hku.hk}
}

\begin{document}
\maketitle

\begin{abstract}
Anticipating financial outcomes requires synthesizing evolving evidence and reasoning under uncertainty, making deep research systems promising tools for forecasting.
Yet evaluating forecasting ability is challenging because historical backtests may reward memorized outcomes or information retrieved after the forecast date.
Live forecasting avoids direct access to realized outcomes, but feedback arrives only after events resolve.
To obtain timely feedback and assess historical scores, we introduce \benchmark{}, a diagnostic backtesting benchmark for financial deep research systems.
Across the Chinese, U.S., and Hong Kong markets, the benchmark evaluates four forecasting tasks: valuation, return ranking, event direction, and future financial fundamentals.
Two capability diagnostic tasks assess counterfactual reasoning and financial calculation, scored separately from forecasting performance.
Our harness-compatible framework restricts retrieved evidence to information available at the forecast date.
We compare pretraining-era, resolved post-cutoff, and live evaluations to test whether controlled backtests approximate future performance.
Pretraining-era scores exceed post-cutoff scores in most temporal comparisons, and stronger retrieval controls narrow the gap between post-cutoff backtests and live results.
Forecasting attribution and robustness analyses reveal how scores respond to retrieval policy and how predictions respond to prompt rewrites and numerical perturbations.
The \href{https://github.com/OpenFinMind/FinForesight}{code} and \href{https://huggingface.co/datasets/OpenFinMind/FinForesight}{benchmark data} are publicly available.
\end{abstract}

\input{01intro}
\input{02related_work}
\input{03methods}
\input{04experiment}
\input{06conclusion}

\clearpage
\input{limitations}

\bibliography{custom}

\input{07appendix}

\end{document}

%% file: 01intro.tex
\section{Introduction}

Financial forecasting requires anticipating how companies and markets will respond to new information.
Such judgments draw on historical prices, corporate disclosures, and news, including evidence that price trends alone cannot capture.
Deep research systems can gather time-sensitive evidence and synthesize it through multi-step reasoning, making them promising for such forecasts \citep{webthinker,mirothinker,fang2026cognitivekernelproframeworkdeep,li-etal-2026-verified}.
In finance, FinRobot integrates financial data and qualitative analysis to produce financial projections and valuation reports \citep{zhou2024finrobot}.
Fin-DR further combines multi-turn evidence retrieval with financial time-series forecasting in a deep research workflow \citep{duan2026findr}.
As these workflows are used for forward-looking analysis, reliably evaluating their predictive ability becomes increasingly important for financial research and investment decisions.

\begin{figure}[t]
\centering
\includegraphics[width=\linewidth]{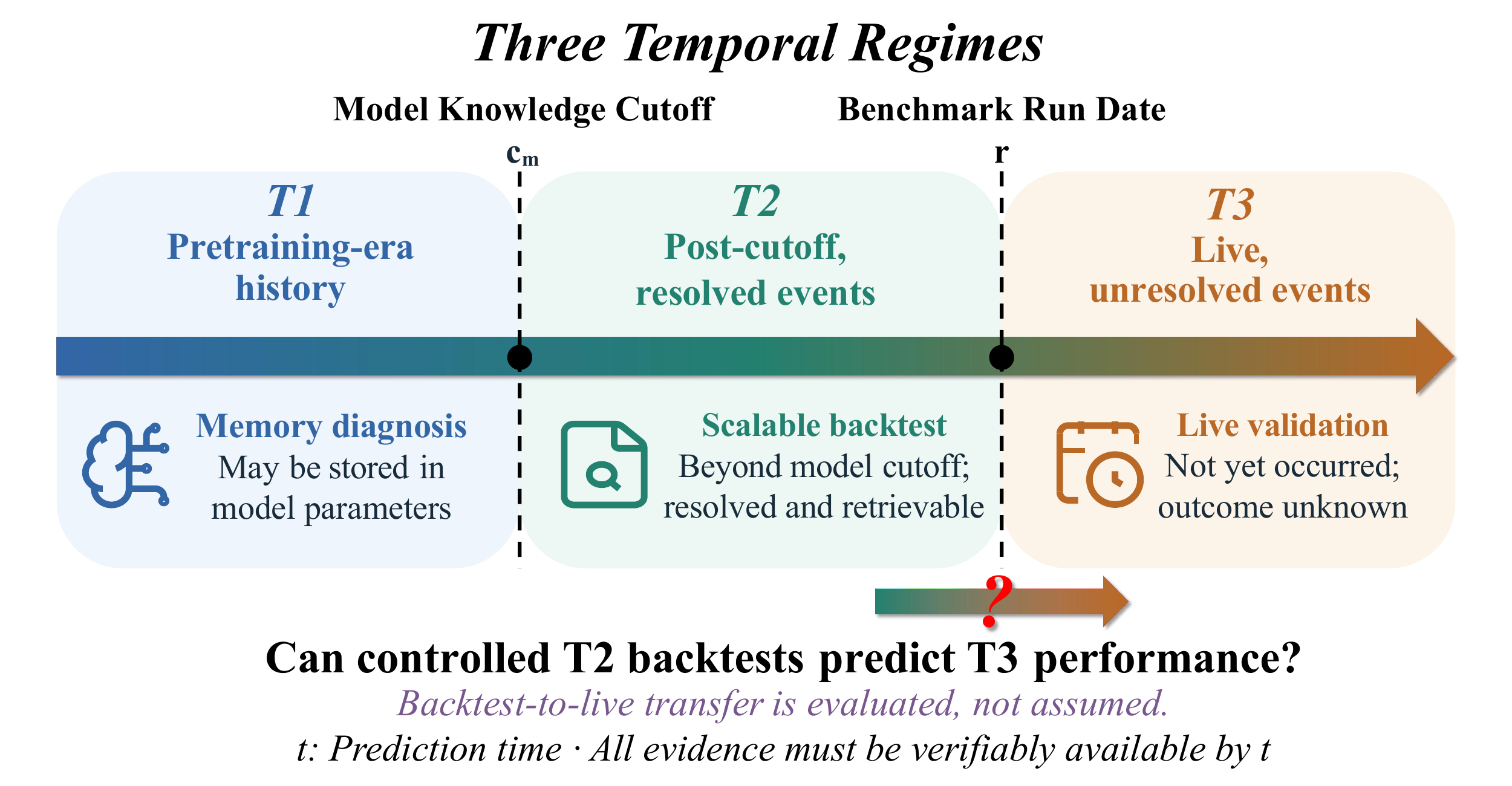}
\vspace{-0.8cm}
\caption{Can controlled T2 backtests approximate T3 forecasting performance?
The model knowledge cutoff $c_m$ marks the T1/T2 boundary, and the benchmark run date $r$ marks the T2/T3 boundary.
T1 contains pretraining-era history, T2 resolved post-cutoff events, and T3 unresolved live events.
Evidence used for each forecast must have been public by the prediction time $t$.}
\label{fig:temporal-regimes}
\vspace{-0.6cm}
\end{figure}

Historical backtests can be scored immediately, but a model may already know the outcome it is asked to predict.
Live evaluation records forecasts before outcomes resolve, but provides feedback only after the events occur.
Figure~\ref{fig:temporal-regimes} places these approaches across three regimes defined by the model knowledge cutoff $c_m$ and benchmark run date $r$.
In T1, historical outcomes may have appeared in training data, allowing \emph{parametric memory} to appear as forecasting ability \citep{li2026simulatedignorance}.
In T2, forecast dates follow the knowledge cutoff and outcomes are resolved by the run date, making scores immediately available without the same pretraining-era exposure.
Yet \emph{retrieval leakage} can expose published outcomes or other evidence unavailable at the original forecast date, even when search date filters are applied \citep{ellahib2026temporalleakage}.
T3 contains live events whose outcomes remain unresolved at the run date and can be scored only after resolution.
This leaves a tradeoff between timely backtest feedback and direct but delayed live validation.

Existing forecasting benchmarks broadly adopt two approaches: live forecasting and pastcasting.
\emph{Live forecasting} records predictions before outcomes are resolved, as in ForecastBench~\citep{karger2024forecastbench}, FutureX~\citep{zeng2025futurex}, and the finance-focused FinDeepForecast~\citep{li2026findeepforecast}.
In contrast, \citet{wildman2025benchtothefuture} introduce Bench to the Future (BTF), a \emph{pastcasting} benchmark that reconstructs historical evidence for resolved questions, enabling immediate evaluation.
This creates a need to compare controlled historical backtests directly with live forecasting outcomes in financial markets.
Interpreting their scores also requires examining historical knowledge, retrieval leakage, and sensitivity to prompt formulations and numerical inputs.
This leads to our central question: \emph{Can Historical Backtests Reliably Evaluate Deep Research for Financial Forecasting?}

To address this question, we introduce \benchmark{}, a diagnostic backtesting benchmark for financial deep research systems.
It spans the Chinese, U.S., and Hong Kong markets and assesses four forecasting targets: stock valuation, relative return ranking, event direction, and future financial fundamentals.
Two capability diagnostic tasks assess counterfactual reasoning and financial calculation, with scores reported separately from the forecasting score.
We generate instances relative to each model's knowledge cutoff and use resolved T2 events to provide timely feedback.
A time-controlled framework supports multiple deep research harnesses and restricts retrieved evidence to information available at the forecast date.
We compare controlled T2 scores with live T3 scores to assess how well historical backtests reflect live forecasting performance.
To interpret these scores, we compare temporal regimes and retrieval policies for signs of memory or information leakage.
Prompt rewrites and numerical perturbations reveal how predictions respond to changes in task wording and numerical inputs.

Our experiments reveal substantial differences across temporal regimes and forecasting tasks.
Pretraining-era evaluations show widespread historical score advantages, and stronger retrieval controls bring T2 backtest scores closer to live results.
Controlled T2 evaluation therefore offers timely feedback for model and system development.
Diagnostic analyses expose sensitivities to information access and input formulation that aggregate forecasting scores alone cannot capture.
Across model--harness configurations, the finance-specific workflow does not consistently outperform general-purpose alternatives.
Together, these findings show why forecasting scores must be interpreted alongside the conditions under which they are obtained.

The main contributions of this work are:
\begin{itemize}[leftmargin=*, itemsep=2pt, topsep=2pt, parsep=0pt]
    \item We build a benchmark spanning four forecasting tasks and two capability diagnostic tasks across the Chinese, U.S., and Hong Kong markets, with resolved T2 events enabling timely evaluation.
    \item We develop a time-controlled evaluation framework compatible with multiple deep research harnesses that generates instances relative to model knowledge cutoffs and audits retrieved evidence against forecast dates.
    \item We analyze forecasting attribution and robustness through temporal comparisons, retrieval-policy tests, prompt rewrites, and numerical perturbations to investigate the sources of apparent forecasting performance.
\end{itemize}

%% file: 02related_work.tex
\section{Related Work}

\subsection{Forecasting Evaluation and Temporal Validity}

Live forecasting records predictions before outcomes resolve but delays evaluation until outcomes become observable \citep{karger2024forecastbench,zeng2025futurex}.
Pastcasting avoids this delay by evaluating resolved questions against reconstructed historical evidence.
Related approaches use frozen retrieval environments for reproducible research-agent evaluation or replay historical prediction markets for retrospective forecasting \citep{bosse2025deepresearchbench,ye2026hindcast}.
Pastcasting studies also compare retrospective evaluations with live forecasting performance \citep{wildman2025benchtothefuture,chai2026agentictime}.
However, whether controlled historical backtests reliably reflect live forecasting performance across financial tasks remains unclear.

Historical forecasting evaluation also faces risks of temporal information leakage.
Models may recall historical financial outcomes, and instructions to disregard future knowledge may fail to prevent their use \citep{didisheim2025predictablememory,li2026simulatedignorance}.
Date-filtered retrieval may also expose information unavailable at the original forecast date, motivating explicit look-ahead audits \citep{ellahib2026temporalleakage,benhenda2026lookaheadbench}.
\benchmark{} combines model-cutoff-aware temporal partitioning with point-in-time retrieval auditing to address these risks in financial deep research evaluation.
It uses T1--T2 comparisons to probe historical performance advantages and T2--T3 comparisons under different retrieval policies to assess backtest-to-live agreement.

\subsection{Financial Agent Benchmarks}

Existing financial-agent benchmarks assess diverse capabilities, including filing-based research, expert investment judgments, and financial decision-making \citep{bigeard2025financeagentbenchmark,cho2026hedgebench,li2024investorbench}.
FinTradeBench~\citep{agrawal2026fintradebench} examines reasoning over fundamental and technical trading signals.
DeepFund~\citep{li2025deepfund} and related benchmarks evaluate investment decisions in market environments \citep{qian2025whenagentstrade,chen2025stockbench}.
Code-based alpha discovery represents another evaluation setting for financial research agents \citep{liu2026cognitivealpha,liu2026xalpha}.
However, aggregate scores across diverse financial tasks may obscure performance on specific future-outcome forecasting tasks.
\benchmark{} therefore evaluates valuation, return ranking, event direction, and future financial fundamentals separately across three markets using task-specific outcome-based scores.

Beyond forecasting accuracy, understanding why systems perform well and whether their predictions are reliable requires complementary diagnostics.
KTD-Fin~\citep{zhu2026ktdfin} investigates the decomposition of trading returns.
Related work examines code validity, numerical reasoning, and prompt-dependent forecasting behavior \citep{luo2026alphaqt,levy2026caution,tang2025financereasoning,schoenegger2025promptforecasting}.
\benchmark{} uses temporal comparisons and retrieval interventions to investigate historical performance advantages and retrieval sensitivity.
Counterfactual reasoning and financial calculation serve as capability diagnostic tasks.
Prompt rewrites and numerical perturbations assess the robustness of forecasting predictions.

%% file: 03methods.tex
\section{Methodology}
\label{sec:methodology}

\subsection{Preliminaries}

\paragraph{Evaluation setup.}
We evaluate the forecasting performance and supporting capabilities of complete \emph{model--harness} systems under controlled information conditions.
Each instance provides a question and task-specific inputs, and the system's response is compared against a reference label accessible only to the evaluator.

\paragraph{Temporal setup.}
Let $c_m$ denote the model's declared knowledge cutoff, $t$ the instance's forecast date, and $r$ the benchmark run date.
In historical backtests, $t$ is the simulated forecast date.
Let $I_t$ denote information verifiably public by $t$.
In time-controlled runs, supplied inputs and retrieved evidence are restricted to $I_t$.
For outcome-based tasks, Figure~\ref{fig:temporal-regimes} distinguishes:
\begin{itemize}[leftmargin=*, itemsep=2pt, topsep=2pt, parsep=0pt]
    \item \textbf{T1: Pretraining-era historical events.}
    Both the forecast date and outcome resolution precede $c_m$, so realized outcomes may have been encountered during pretraining.

    \item \textbf{T2: Post-cutoff resolved events.}
    The forecast date follows $c_m$, and outcomes resolve between $t$ and $r$, enabling immediate scoring.

    \item \textbf{T3: Live future events.}
    Predictions are recorded before outcomes resolve, with reference labels becoming available only after $r$.
\end{itemize}

\paragraph{Design principles.}
We adopt four principles to support forecasting evaluation, attribution, and robustness analysis.
\begin{itemize}[leftmargin=*, itemsep=2pt, topsep=2pt, parsep=0pt]
    \item \textbf{Temporal control beyond prompts.}
    We enforce point-in-time information access and use temporal comparisons to probe potential historical knowledge effects \citep{li2026simulatedignorance}.

    \item \textbf{Page-level retrieval audits.}
    We audit retrieved content against the forecast date, since timestamps and search filters alone cannot guarantee temporal validity \citep{ellahib2026temporalleakage}.

    \item \textbf{Decomposed evaluation.}
    We report performance separately by task, market, and temporal regime to distinguish forecasting capabilities and support diagnostic comparisons.

    \item \textbf{Response robustness.}
    We use prompt rewrites to assess prediction consistency and numerical perturbations to examine sensitivity to input changes \citep{schoenegger2025promptforecasting,levy2026caution}.
\end{itemize}

\subsection{Benchmark}

\begin{figure*}[t]
\centering
\includegraphics[width=\linewidth]{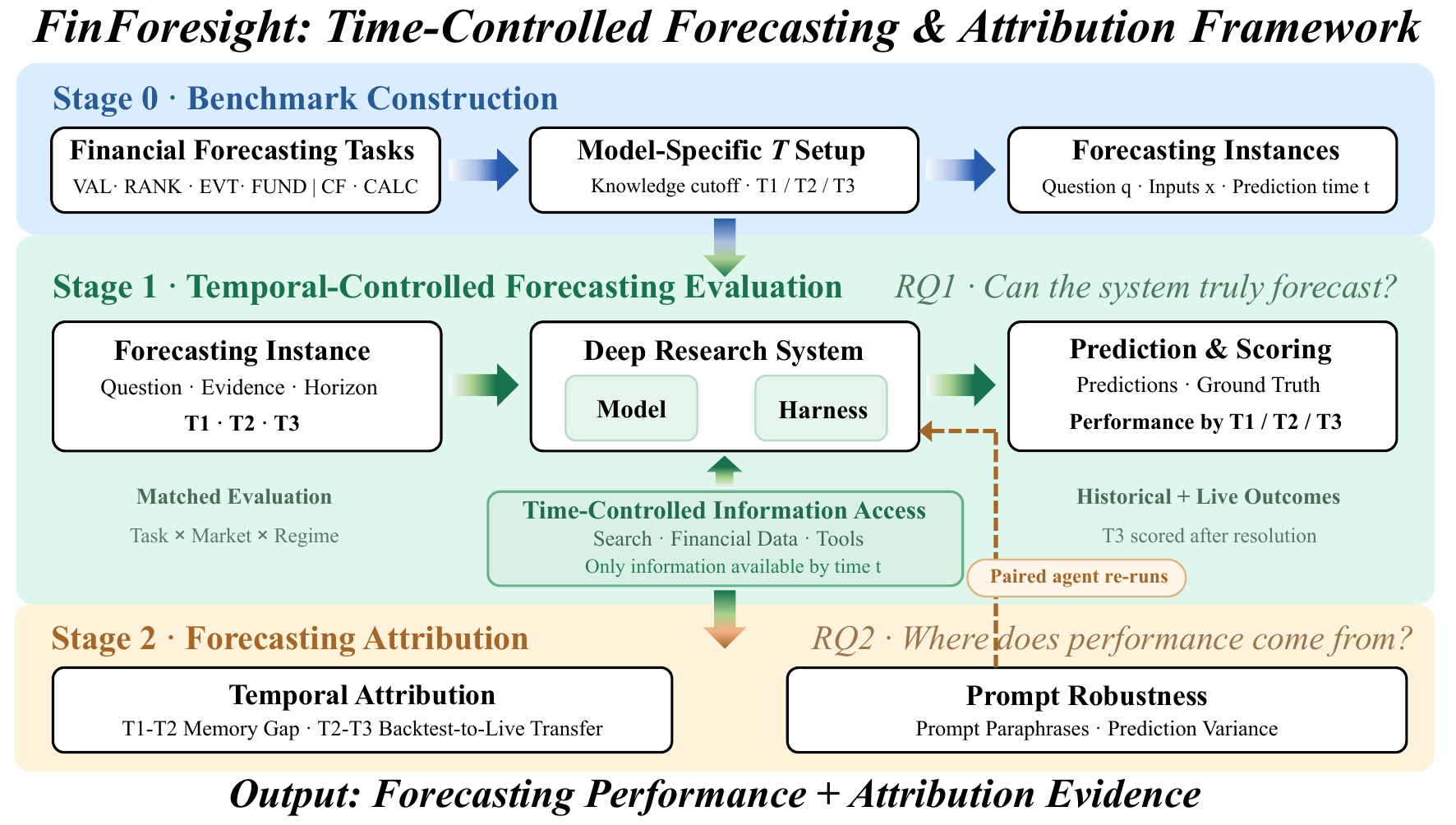}
\vspace{-0.8cm}
\caption{Three-stage workflow of \benchmark{}.
Model-specific instances are constructed across financial tasks, markets, and T1/T2/T3 temporal regimes.
The model--harness system generates predictions using only external information available at the forecast date.
The evaluator scores predictions against reference labels immediately for resolved outcomes or after resolution for live events.
Temporal comparisons assess historical performance advantages and backtest-to-live agreement, and prompt perturbations examine prediction robustness.}
\label{fig:backtesting-framework}
\vspace{-0.6cm}
\end{figure*}

\paragraph{Benchmark construction.}
We construct \benchmark{} dynamically across six task families and the Chinese, U.S., and Hong Kong markets, adapting instances to each model's knowledge cutoff.
Task templates generate questions, required inputs, and answer schemas, with target dates and prediction windows constrained by the applicable temporal conditions.
We retain candidates with sufficient inputs and reproducible reference-label rules.
We verify evidence timestamps and confirm that the information was publicly available by $t$.
Forecasting labels are derived from subsequently realized market or financial outcomes, whereas capability diagnostic answers follow predefined logical or computational rules.
T3 labels are added after the corresponding outcomes resolve.
Market prices are obtained from Yahoo Finance \citep{yahoofinancehistorical}.
Filings and financial values are collected from SEC EDGAR, CNINFO, the Shanghai and Shenzhen stock exchanges, and HKEXnews \citep{secedgar,cninfodisclosures,ssedisclosures,szsedisclosures,hkexnewsdisclosures}.
Macroeconomic events are sourced from releases by the U.S. Bureau of Labor Statistics, China's National Bureau of Statistics, and Hong Kong's Census and Statistics Department \citep{blsreleases,nbsreleases,censtatdreleases}.
We compare systems on matched eligible instances to improve comparability across different model knowledge cutoffs.
Source availability at the forecast date is verified independently of outcome-label collection.

\paragraph{Task suite.}
Equity research involves assessing stock valuations, relative returns among industry peers, market reactions to events, and future financial performance.
\benchmark{} evaluates these judgments through four forecasting tasks: Valuation (VAL), Return Ranking (RANK), Event Direction (EVT), and Fundamentals (FUND).
Their reference labels are derived from realized outcomes, and their scores contribute to the overall forecasting score.
Two capability diagnostic tasks, Counterfactual (CF) and Calculation (CALC), assess conditional reasoning and financial calculation, respectively.
Their reference answers are predefined, and their scores are reported separately from the overall forecasting score.
Appendix~\ref{app:task-templates} presents each task's input, output, scoring rule, and example.
\begin{itemize}[leftmargin=*, itemsep=2pt, topsep=2pt, parsep=0pt]

\item \textbf{Valuation (VAL).}
VAL forecasts a stock's future price through bear, base, and bull targets, together with an expected time bucket for reaching the predicted range.
Predictions use point-in-time evidence and are evaluated against subsequent market prices.

\item \textbf{Return Ranking (RANK).}
RANK evaluates relative return forecasting by ranking industry peers according to their expected excess returns over a specified horizon.
It includes fundamental-only, technical-only, and hybrid variants, with rankings compared against realized returns.

\item \textbf{Event Direction (EVT).}
EVT predicts market reactions to earnings, policy, and macroeconomic events using information available at the forecast date.
The system outputs a return direction and upward probability, evaluated against realized market movements.

\item \textbf{Fundamentals (FUND).}
FUND forecasts company financial metrics that have not yet been disclosed at the prediction date.
Predicted values are compared with subsequent financial disclosures.

\item \textbf{Counterfactual (CF).}
CF assesses conditional reasoning through paired fictional announcements with contrasting financial implications.
It tests whether predicted excess-return directions follow the supplied event conditions, including cases that conflict with historical price trends.

\item \textbf{Calculation (CALC).}
CALC evaluates multi-step financial numerical reasoning using problems with distractor values and unit or convention traps.
Intermediate calculations and final answers are checked against predefined references.

\end{itemize}

\paragraph{Task example.}
The following example presents an abbreviated VAL task instance, its recorded price targets, and the held-out 30-day reference price.
Additional task examples appear in Appendix~\ref{app:task-templates}, where missing diagnostic records are highlighted in red.

\begin{center}
\begingroup
\setlength{\fboxsep}{6pt}
\fbox{\begin{minipage}{\dimexpr\columnwidth-2\fboxsep-2\fboxrule\relax}
\small
\textbf{VAL: Wuliangye (000858), CN.}

\textbf{Question.}
Using public information available before 2026-01-30, estimate bear, base, and bull price targets over a 30-day forecast horizon.
The price at the forecast date is CNY 105.00 per share.

\textbf{Recorded price targets.}
GPT-4.1-mini with GPT-Researcher, using search, PIT control, and the LLM filter:
\begin{center}
\footnotesize
\begin{tabular*}{\linewidth}{@{\extracolsep{\fill}}lr@{}}
\texttt{bear} & CNY 95.00 \\
\texttt{base} & CNY 110.00 \\
\texttt{bull} & CNY 125.00 \\
\end{tabular*}
\end{center}

\textbf{Held-out reference.}
The observed 30-day reference price is CNY 104.05 per share on 2026-02-27.
\end{minipage}}
\endgroup
\end{center}

\subsection{Time-Controlled Backtesting Framework}
\label{sec:backtesting-framework}

Figure~\ref{fig:backtesting-framework} illustrates the three-stage workflow used by \benchmark{} to evaluate model--harness systems, comprising benchmark construction, forecasting evaluation, and prediction attribution.
For each eligible instance, the controller configures the specified model, harness, and retrieval policy before supplying task inputs.
Reference labels remain accessible only to the evaluator.
The harness accesses search, market data, and other tools through mediated services, with all interactions logged for subsequent evidence audits.
We evaluate three retrieval policies: search without temporal checks, point-in-time (PIT) control, and PIT control augmented with an LLM-based leakage filter.
System responses are recorded and locked before scoring.
Resolved T1/T2 forecasts and capability diagnostic tasks are scored immediately, while T3 forecasts are scored after their outcomes resolve.
Task scores follow the protocol in Section~\ref{sec:experiments}.
Temporal comparisons and retrieval-policy interventions provide diagnostic evidence about historical performance advantages and the influence of external information.
Prompt rewrites and numerical perturbations further examine prediction robustness.

\paragraph{LLM leakage filter.}
\label{sec:llm-filter}
PIT control uses source timestamps and available historical versions to restrict retrieved evidence to $I_t$, rejecting pages published after the forecast date $t$.
We additionally use DeepSeek-V4-Flash~\citep{deepseek2026v4flash}, specifically its April 2026 release, as an LLM-based leakage filter over retrieved snippets and page content.
The filter applies a fixed classification prompt and decoding configuration across all runs to determine whether retrieved content was publicly available at $t$.
Content containing post-$t$ information, ambiguous temporal evidence, uncertain classifications, or invalid filter responses is rejected before reaching the evaluated system.

%% file: 04experiment.tex
\section{Experiments}
\label{sec:experiments}

\subsection{Experimental Settings}

\begin{table*}[t]
\centering
\small
\setlength{\tabcolsep}{4pt}
\resizebox{\textwidth}{!}{%
\begin{tabular}{llrrrrrrr}
\toprule
\rowcolor{tableHeader}
 & & \multicolumn{5}{c}{\textbf{Forecasting Performance}} & \multicolumn{2}{c}{\textbf{Capability Diagnostic Tasks}} \\
\cmidrule(lr){3-7}\cmidrule(lr){8-9}
\rowcolor{tableHeader}
\textbf{Harness} & \textbf{Model}
& \textbf{VAL$\uparrow$}
& \textbf{RANK$\uparrow$}
& \textbf{EVT$\uparrow$}
& \textbf{FUND$\uparrow$}
& \textbf{Overall$\uparrow$}
& \textbf{CF$\uparrow$}
& \textbf{CALC$\uparrow$} \\
\midrule
\multirow{6}{*}{FinanceHarness}
& Gemma-3
& 0.7971 & 0.4884 & 0.6722 & 0.4061 & 0.5910 & 0.8667 & 0.6199 \\

& GPT-4.1-mini
& 0.8746 & 0.4687 & 0.6491 & 0.6105 & 0.6507 & 0.9000 & 0.5517 \\

& Phi-4
& 0.8857 & 0.4536 & 0.6774 & 0.6493 & 0.6665 & 0.9000 & 0.8005 \\

& Claude Sonnet 4.5
& 0.8706 & 0.4785 & 0.6717 & 0.6979 & 0.6797 & 0.9667 & 0.8519 \\

& DeepSeek V4 Flash
& 0.8576 & 0.4869 & 0.6765 & 0.7337 & \underline{0.6887} & 0.9667 & 1.0000 \\

& Gemini 3.1 Flash-Lite
& 0.8578 & 0.4960 & 0.7183 & 0.7476 & \textbf{0.7049} & 1.0000 & 0.8929 \\

\midrule

\multirow{6}{*}{Open Deep Research}
& GPT-4.1-mini
& 0.6212 & 0.4547 & 0.6402 & 0.6910 & 0.6018 & 0.9333 & 0.3182 \\

& Gemma-3
& 0.7153 & 0.4699 & 0.6804 & 0.5698 & 0.6089 & 0.9000 & 0.3177 \\

& Claude Sonnet 4.5
& 0.8696 & 0.4428 & 0.6275 & 0.7830 & 0.6807 & 0.9667 & 0.6520 \\

& Phi-4
& 0.8956 & 0.4648 & 0.6640 & 0.7049 & 0.6823 & 0.9333 & 0.5338 \\

& DeepSeek V4 Flash
& 0.7905 & 0.5023 & 0.6581 & 0.7798 & \underline{0.6827} & 0.9333 & 0.9333 \\

& Gemini 3.1 Flash-Lite
& 0.7468 & 0.5022 & 0.7401 & 0.7706 & \textbf{0.6899} & 1.0000 & 0.5000 \\

\midrule

\multirow{6}{*}{GPT-Researcher}
& Gemma-3
& 0.7421 & 0.4676 & 0.7013 & 0.3301 & 0.5603 & 0.8667 & 0.3262 \\

& Phi-4
& 0.8635 & 0.4659 & 0.7032 & 0.4466 & 0.6198 & 0.9000 & 0.5436 \\

& Claude Sonnet 4.5
& 0.7833 & 0.4648 & 0.6524 & 0.7077 & 0.6521 & 0.9655 & 0.7143 \\

& GPT-4.1-mini
& 0.7840 & 0.4673 & 0.6998 & 0.6751 & 0.6566 & 0.9333 & 0.2143 \\

& DeepSeek V4 Flash
& 0.8384 & 0.4910 & 0.6855 & 0.7974 & \underline{0.7031} & 0.9667 & 1.0000 \\

& Gemini 3.1 Flash-Lite
& 0.8420 & 0.4697 & 0.7502 & 0.7778 & \textbf{0.7099} & 0.9667 & 0.2143 \\

\midrule

\multirow{2}{*}{Claude Code}
& Claude Sonnet 4.5
& 0.6793 & 0.4690 & 0.8108 & 0.5578 & \underline{0.6292} & 1.0000 & 1.0000 \\

& Claude Opus 4.6
& 0.7596 & 0.4554 & 0.7550 & 0.6202 & \textbf{0.6476} & 1.0000 & 1.0000 \\

\midrule

\multirow{2}{*}{Codex}
& GPT-4.1-mini
& 0.6302 & 0.4685 & 0.7480 & 0.6287 & \underline{0.6189} & 1.0000 & 0.5000 \\

& GPT-5.4
& 0.7907 & 0.4636 & 0.7809 & 0.6359 & \textbf{0.6678} & 1.0000 & 0.6667 \\
\bottomrule
\end{tabular}%
}
\vspace{-0.2cm}
\caption{Forecasting performance on T2 instances and capability diagnostic task scores across 22 model--harness configurations.
Overall is the unweighted mean of VAL, RANK, EVT, and FUND, excluding CF and CALC.
Rows are ordered by increasing Overall within each harness.
Bold and underlined Overall scores indicate the highest- and second-highest-scoring configurations within each harness, respectively.}
\label{tab:main-forecasting-t2}
\vspace{-0.4cm}
\end{table*}

\paragraph{Models and Harnesses.}
We evaluate three types of harnesses: financial research, general research, and coding agents.
\textsc{FinanceHarness}~\cite{xiao2026financeharness} specializes in financial research.
\textsc{GPT-Researcher}~\cite{elovic2026gptresearcher} and \textsc{Open Deep Research}\footnote{\url{https://github.com/langchain-ai/open_deep_research}} support general-domain research, with the latter built on LangGraph~\cite{langgraph2024}.
\textsc{Claude Code}\footnote{\url{https://claude.com/product/claude-code}} and \textsc{Codex}~\cite{openai2026codex} are general-purpose coding agents.
Coding agents receive the same task instructions and access to the same research tools as the research-focused harnesses.
Across 22 model--harness configurations, we evaluate eight large language models: GPT-4.1-mini, Claude Sonnet 4.5, Gemini 3.1 Flash-Lite, Gemma 3~\cite{gemmateam2025gemma3}, Phi-4~\cite{abdin2024phi4}, DeepSeek-V4-Flash~\cite{deepseek2026v4flash}, Claude Opus 4.6, and GPT-5.4~\cite{openai2026gpt54}.
This selection prioritizes temporal comparability over model recency: all systems share a T2 window and are scored on resolved outcomes.
Evaluating newer models with later knowledge cutoffs would require moving this window forward and waiting for outcomes to resolve.
The configurable dataset generators in \benchmark{} support constructing new shared T2 windows for future frontier models.

\paragraph{Data and Tasks.}
We construct \benchmark{} across the Chinese, U.S., and Hong Kong markets using configurable task generators and publicly available financial and market data.
Instances are assigned to T1, T2, or T3 according to each model's knowledge cutoff and the temporal definitions in Section~\ref{sec:methodology}.
The dataset uses forecast dates before 2024-06-01 for T1 and after 2025-07-31 for T2, subject to each model's knowledge cutoff.
T2 serves as the primary evaluation setting.
T1 and T3 support temporal comparisons.
The suite contains 1,119 records: VAL (244), RANK (383), EVT (138), FUND (294), CF (30), and CALC (30).
The four forecasting tasks evaluate future financial outcomes.
The two capability diagnostic tasks, CF and CALC, assess counterfactual reasoning and financial calculation.
Task-specific forecast horizons, outcome resolution rules, and scoring criteria are detailed in Appendix~\ref{app:task-templates}.

\paragraph{Evaluation Metrics.}
VAL measures valuation error reduction relative to the cutoff-price baseline and reversion-horizon accuracy.
RANK evaluates future excess-return rankings using normalized Spearman correlation and top-$K$ hit rate.
EVT evaluates probabilistic directional forecasts using Brier skill, and FUND measures numerical prediction accuracy through a clipped per-instance utility.
CF measures adherence to injected event logic, with paired-response consistency and trend-conflict adherence reported separately.
CALC averages tolerance-based Exact Match over intermediate and final numerical steps, then across problems.
Forecasting attribution uses Delta(T1,T2) and Delta(T2,T3), the signed score differences between temporal regimes, alongside retrieval-policy comparisons.
Prompt Change Rate and Perturb Change Rate measure material output changes across valid prompt and numerical comparison pairs, respectively.
Score Delta averages the signed score changes induced by prompt rewrites and numerical perturbations.
Appendix~\ref{app:task-templates} provides task-specific scoring formulas, and Appendix~\ref{app:metric-formulas} defines the temporal and robustness diagnostics.

\begin{table}[t]
\centering
\footnotesize
\setlength{\tabcolsep}{3pt}
\renewcommand{\arraystretch}{1.15}
\begin{tabularx}{\columnwidth}{Xlrrr}
\toprule
\rowcolor{tableHeader}
\textbf{Model} & \textbf{Task} & \textbf{CN} & \textbf{US} & \textbf{HK} \\
\midrule
\multirow{3}{*}{GPT-4.1-mini} & RANK & +0.0564 & +0.0635 & +0.0346 \\
 & EVT & +0.0549 & +0.0599 & -0.0117 \\
 & FUND & +0.4469 & +0.1045 & +0.1602 \\
\midrule
\multirow{3}{*}{\shortstack[l]{Gemini 3.1\\Flash-Lite}} & RANK & +0.0906 & +0.1709 & +0.1555 \\
 & EVT & +0.1013 & +0.0259 & -0.0079 \\
 & FUND & +0.3503 & +0.0897 & +0.1523 \\
\midrule
\multirow{3}{*}{\shortstack[l]{Claude Sonnet\\4.5}} & RANK & +0.1135 & +0.0765 & +0.0992 \\
 & EVT & +0.1227 & +0.1108 & +0.0482 \\
 & FUND & +0.4341 & +0.0859 & +0.2182 \\
\bottomrule
\end{tabularx}
\vspace{-0.2cm}
\caption{T1--T2 score gaps for RANK, EVT, and FUND across CN/US/HK in FinanceHarness.
Each cell reports $\Delta(T1,T2)=S_{T1}-S_{T2}$.
Positive values indicate higher T1 scores than T2 scores.}
\label{tab:temporal-summary}
\vspace{-0.4cm}
\end{table}

\subsection{Main Forecasting Benchmark}

Table~\ref{tab:main-forecasting-t2} compares 22 model--harness configurations on T2.
Overall is the mean of VAL, RANK, EVT, and FUND.
CF and CALC are reported separately as capability diagnostics.

\paragraph{Forecasting strengths vary across tasks.}
Gemini 3.1 Flash-Lite achieves the highest Overall and EVT scores across all three research harnesses, while Phi-4 consistently leads VAL.
These differences between valuation and event-direction results show that aggregate rankings do not fully capture task-specific forecasting strengths, making model selection dependent on the forecasting objective.

\paragraph{Harness effects and forecasting capabilities.}
FinanceHarness achieves the highest mean Overall score across the six shared models, but no single research harness consistently dominates.
For example, DeepSeek V4 Flash and Gemini 3.1 Flash-Lite perform best with GPT-Researcher, demonstrating that harness advantages depend on the underlying model.
Moreover, Claude Sonnet 4.5 achieves perfect CF and CALC scores with Claude Code but lower Overall forecasting performance than with FinanceHarness.
These findings highlight the effect of harness selection on forecasting performance.
They also show that strong reasoning and calculation capabilities do not necessarily translate into better future-outcome predictions.

\subsection{Temporal Regime Analysis}

We compare GPT-4.1-mini, Gemini 3.1 Flash-Lite, and Claude Sonnet 4.5 in FinanceHarness on RANK, EVT, and FUND forecasting tasks.
VAL is excluded from this focused analysis due to its sensitivity to short-horizon price fluctuations, with its temporal results reported in Table~\ref{tab:app-main-full-matrix} in Appendix~\ref{app:experiment-details}.

\begin{table}[t]
\centering
\footnotesize
\setlength{\tabcolsep}{3pt}
\renewcommand{\arraystretch}{1.15}
\begin{tabularx}{\columnwidth}{Xlrrr}
\toprule
\rowcolor{tableHeader}
\textbf{Model} & \shortstack[l]{\textbf{Search Setting}} & \textbf{RANK} & \textbf{EVT} & \textbf{FUND} \\
\midrule
\multirow{3}{*}{\shortstack[l]{GPT-4.1\\mini}} & Search & +0.0132 & +0.0659 & +0.1337 \\
 & + PIT & +0.0093 & +0.0684 & +0.0328 \\
 & + PIT + Filter & \textbf{+0.0037} & \textbf{+0.0043} & \textbf{+0.0141} \\
\midrule
\multirow{3}{*}{\shortstack[l]{Gemini 3.1\\Flash-Lite}} & Search & +0.1286 & +0.1077 & +0.1465 \\
 & + PIT & \textbf{+0.0748} & +0.0161 & +0.0473 \\
 & + PIT + Filter & +0.0864 & \textbf{+0.0103} & \textbf{+0.0344} \\
\midrule
\multirow{3}{*}{\shortstack[l]{Claude\\Sonnet 4.5}} & Search & +0.1246 & +0.2306 & +0.1320 \\
 & + PIT & \textbf{+0.0537} & +0.1882 & +0.0659 \\
 & + PIT + Filter & +0.0689 & \textbf{+0.0080} & \textbf{+0.0267} \\
\bottomrule
\end{tabularx}
\vspace{-0.2cm}
\caption{US T2--T3 score gaps across three retrieval settings in FinanceHarness, using matched one-week evaluation windows.
Gaps are computed as T2 scores minus T3 scores, with positive values indicating higher T2 performance.
Bold indicates the smallest absolute gap for each model--task pair.}
\label{tab:temporal-summary2}
\vspace{-0.4cm}
\end{table}

\paragraph{Historical advantages are widespread.}
T1 scores exceed T2 scores in 25 of 27 comparisons across the three markets, including all RANK and FUND comparisons (Table~\ref{tab:temporal-summary}).
The advantage appears across all three models, suggesting that each has memorized some historical outcomes \citep{didisheim2025predictablememory}.
Pretraining-era backtests can therefore make recall look like prediction and should not be used alone to assess forecasting ability.
Full scores appear in Table~\ref{tab:app-main-full-matrix} in Appendix~\ref{app:experiment-details}.

\paragraph{Temporal controls improve backtest-to-live agreement.}
We compare T2 and T3 scores over matched one-week windows under search only, point-in-time (PIT) control, and PIT plus an LLM leakage filter (Table~\ref{tab:temporal-summary2}).
With search alone, the mean absolute T2--T3 gap across nine model--task pairs is 0.1203.
PIT lowers the mean gap to 0.0618, leaving a gap that may reflect pages updated after the forecast date despite their earlier displayed timestamps \citep{ellahib2026temporalleakage}.
Adding the LLM filter reduces the mean gap to 0.0285, approximately 76\% below search alone, with smaller gaps in all nine comparisons.
Under this stricter control, post-cutoff backtests closely approximate live forecasting scores in the evaluated tasks.
Because T2 outcomes are already resolved, they provide timely feedback for model and system development.
Full scores appear in Table~\ref{tab:app-filter-transfer}.

\begin{table*}[t]
\centering
\footnotesize
\setlength{\tabcolsep}{5pt}
\renewcommand{\arraystretch}{1.12}
\begin{tabularx}{\textwidth}{lc*{4}{>{\centering\arraybackslash}X}}
\toprule
\rowcolor{tableHeader}
\textbf{Model} & \textbf{Task} & {\boldmath$\Delta(T1,T2)$} & {\boldmath$\Delta(T2,T3)$} & \textbf{T3 Score} & \shortstack{\textbf{Input Score}\\\textbf{Delta}} \\
\midrule
\multirow{2}{*}{GPT-4.1-mini} & EVT & +0.0599 & +0.0043 & 0.5805 & -0.003848 \\
 & FUND & +0.1045 & +0.0141 & 0.7194 & -0.001075 \\
\midrule
\multirow{2}{*}{\shortstack[l]{Claude Sonnet\\4.5}} & EVT & +0.1108 & +0.0080 & 0.5926 & +0.000235 \\
 & FUND & +0.0859 & +0.0267 & 0.8563 & -0.000910 \\
\bottomrule
\end{tabularx}
\vspace{-0.2cm}
\caption{Prediction attribution for US EVT and FUND in FinanceHarness.
$\Delta(T1,T2)=S_{T1}-S_{T2}$ and $\Delta(T2,T3)=S_{T2}-S_{T3}$ use T2 scores under search, PIT, and LLM filtering.
Input Score Delta combines signed score changes under prompt rewrites and numerical perturbations.}
\label{tab:attribution-summary}
\vspace{-0.4cm}
\end{table*}

\subsection{Market-Level Analysis}

Table~\ref{tab:market-summary} in Appendix~\ref{app:market-level-analysis} compares model performance across CN, US, and HK, averaging scores equally across the three research harnesses within each market.
Table~\ref{tab:app-market-details} in the same appendix section reports the underlying model--harness scores.
All six models score highest in the US, but the market leader changes across CN, US, and HK.
DeepSeek V4 Flash narrowly leads in CN (0.6825), and Gemini 3.1 Flash-Lite leads in US (0.7480) and HK (0.6698).
Gemini remains second in CN, and DeepSeek ranks among the top three in US and HK, so their strengths extend beyond a single market.
No model leads in all three markets, making market-specific scores important alongside the pooled Overall score.

\subsection{Robustness and Sensitivity}

We assess input robustness on T2 EVT and FUND across CN, US, and HK using GPT-4.1-mini and Claude Sonnet 4.5 with FinanceHarness and GPT-Researcher.
Prompt variants rephrase the same question without changing its meaning or available evidence.
Numerical variants slightly alter decimal precision, units, or nonessential values, with equivalent units accepted in scoring.
Prompt Change Rate and Perturb Change Rate measure material changes in predictions, and Score Delta summarizes the mean signed score change across both interventions.
Appendix~\ref{app:experiment-details} reports aggregate results in Table~\ref{tab:robustness-summary} and task--market results in Table~\ref{tab:app-robustness-observed}.
Across the four model--harness configurations, Prompt Change Rates range from 16.83\% to 23.67\%, exceeding Perturb Change Rates of 7.00\% to 9.17\%.
Both prompt wording and small numerical changes can alter individual predictions, with wording having the larger effect on outputs.
Yet the mean signed Score Delta stays below 0.0026 in magnitude for every configuration, so these output changes have little effect on aggregate scores.
This score stability supports task understanding across controlled input variations.

\subsection{Prediction Attribution}

Table~\ref{tab:attribution-summary} combines temporal performance and input sensitivity for GPT-4.1-mini and Claude Sonnet 4.5 on US EVT and FUND in FinanceHarness.
Both temporal gaps use T2 scores with search, point-in-time (PIT) control, and LLM filtering.
The T3 column reports live forecasting scores directly.
The final column reports the mean signed score change under prompt rewrites and numerical perturbations.

All four T1 scores exceed their controlled T2 counterparts, suggesting an advantage from historical knowledge.
Both models retain meaningful performance on live outcomes, with T3 scores of 0.5805--0.5926 on EVT and 0.7194--0.8563 on FUND.
These scores indicate forecasting ability on events whose outcomes were unknown when the predictions were made.
Filtered T2--T3 gaps range from 0.0043 to 0.0267, supporting controlled T2 as a close proxy for live scores in these four comparisons.
Input Score Deltas remain below 0.004 in magnitude across all four comparisons, indicating small average score changes under prompt and numerical variants.
Together, live performance, small filtered T2--T3 gaps, and stable aggregate scores provide evidence that these models understand the forecasting tasks.
The retrieval-policy scores appear in Appendix~\ref{app:prediction-attribution}.

%% file: 06conclusion.tex
\section{Conclusion}

We introduced \benchmark{} with four forecasting and two capability diagnostic tasks across three markets, using a time-controlled, multi-harness framework.
Strong capability scores do not ensure stronger forecasts, and T1's advantage over T2 points to historical knowledge effects.
Model strengths vary across tasks and markets, and the finance-specific harness does not consistently outperform general-purpose workflows.
In the matched one-week US experiment, PIT and filtering bring T2 backtests closer to live T3 scores, providing timely development feedback.
Prompt rewrites and numerical perturbations alter some individual predictions, but have little effect on aggregate scores.
For future work, we will test whether fresh T2 feedback helps autoresearch systems improve T3 forecasts, probing recursive self-improvement (RSI) in finance.

%% file: limitations.tex
\section*{Limitations}

The benchmark has several limitations. First, point-in-time reconstruction is only an approximation: web pages, metadata, snippets, and archived sources may not perfectly match what an investor or deployed agent could access at the cutoff date. Second, CN, US, and HK markets differ in disclosure rules, language coverage, data vendors, and information-release timing, so cross-market results should be interpreted as diagnostic stability evidence rather than clean causal estimates of market difficulty. Third, T2 is a practical retrospective proxy, not a substitute for T3 live evaluation; the T2--T3 relationship must be measured empirically. 

\section*{Ethical Considerations}

AI assistance in manuscript preparation was limited to grammar correction and refinement of logical clarity.
All edits were reviewed by the authors, who take responsibility for the final manuscript.

%% file: 07appendix.tex
\clearpage
\appendix

\section{Tasks, Scoring, and Examples}
\label{app:task-templates}

This appendix presents each task's motivation, full definition, input and output requirements, scoring rule, and a compact execution example.
The examples pair system predictions with reference values used by the evaluator.
Dated forecasting instances use evidence available by their cutoff, while reference outcomes are collected after resolution.
Task prompts request structured outputs, and reference labels remain evaluator-only.
VAL evaluates 30-, 90-, 180-, and 365-day windows as they resolve, RANK uses a 30-day window in the examples below, and EVT and FUND use the window or reporting period specified by each question.
CF and CALC use predefined logic and numerical reference answers, not future market outcomes.
The pooled Overall Score therefore reflects only the four outcome-based forecasting tasks.

\subsection{Valuation (VAL)}

Equity research often culminates in a valuation judgment, as illustrated by FinRobot's research-and-valuation workflow~\citep{zhou2024finrobot}.
VAL turns this scenario-based judgment into an outcome-based price forecast.
An instance specifies a stock, cutoff date and closing price, admissible evidence, and target windows.
The system returns bear, base, and bull prices and a \texttt{reversion\_horizon}.
The base-price forecast is compared with the realized price at each settled horizon, and the horizon prediction is compared with the first observed window in which the price enters the bear--bull range.
Using the cutoff price as a baseline, the relative price skill and VAL score are
\begin{align}
S_{\mathrm{rel}} &= 1-\frac{\mathrm{err}(\hat p_{\mathrm{base}},p)}
                         {\mathrm{err}(p_{\mathrm{cutoff}},p)}, \\
s_{\mathrm{VAL},c} &= \bigl(g(S_{\mathrm{rel}})+H_h\bigr)/2,
\end{align}
where $H_h$ is the reversion-horizon hit rate and $g(z)=\min(1,\max(0,z))$.
Errors are aggregated over the evaluated scope before forming the skill ratio.
The score gives equal weight to improvement over the cutoff-price baseline and correct timing of entry into the forecast range.
Range hit, $\mathbf{1}[\hat p_{\mathrm{bear}}\le p\le\hat p_{\mathrm{bull}}]$, is reported separately to show interval coverage.

\textbf{Example.}\par
\textit{Question:} At the 2026-01-30 cutoff, Wuliangye closed at CNY 105.0; predict bear, base, and bull prices and when the price will enter that range.\par
\textit{Prediction:} Bear/base/bull prices are CNY 95/110/125, with a 3--6-month reversion horizon.\par
\textit{Reference:} The realized 30-day closing price is CNY 104.05, and the 365-day outcome was unresolved for this evaluation.\par
\textit{Scoring:} The 30-day price falls inside the predicted range, but the predicted reversion horizon differs from the first observed entry window.

\subsection{Return Ranking (RANK)}

RANK reflects the analyst's choice among industry peers and follows the distinction between fundamental and trading signals in FinTradeBench~\citep{agrawal2026fintradebench}.
A peer ranking tests relative return forecasts that a single-stock price target cannot capture.
It supplies an industry, a stock cohort, a cutoff, and a forecast window.
RANK-F provides fundamental signals, RANK-T provides technical signals, and RANK-H provides both.
The system must return a complete ranking without duplicate or missing stock codes.
Realized benchmark-relative returns determine the reference ordering.
Normalized Spearman correlation measures agreement across the full ranking, and Top-$K$ Hit measures overlap among the highest-ranked stocks:
\begin{align}
\rho &= 1-\frac{6\sum_i(r_i^{\mathrm{pred}}-r_i^{\mathrm{act}})^2}{N(N^2-1)}, \\
\mathrm{TopKHit} &= \frac{|\mathrm{TopK}_{\mathrm{pred}}\cap\mathrm{TopK}_{\mathrm{act}}|}{K}, \\
s_{\mathrm{RANK},c} &= \bigl((\rho+1)/2+\mathrm{TopKHit}\bigr)/2.
\end{align}
Here $K=\lfloor N/3\rfloor$ and $(\rho+1)/2$ maps Spearman correlation to $[0,1]$.
Higher scores indicate better identification of relative winners across the peer group.
The two terms reward both the complete ordering and the selection of top-performing stocks.
Long--short return spread is an auxiliary measure.

\textbf{Example.}\par
\textit{Question:} Rank 12 Chinese consumer stocks by expected 30-day return using fundamental signals available on 2025-09-30.\par
\textit{Prediction:} The predicted ranking begins with \texttt{000651}, \texttt{000333}, \texttt{002714}, and \texttt{000568}.\par
\textit{Reference:} The realized top four are \texttt{002304}, \texttt{000333}, \texttt{601888}, and \texttt{000651}.\par
\textit{Scoring:} Two predicted top-four stocks enter the realized top four, so $\mathrm{TopKHit}=2/4$; Spearman correlation uses all 12 ranks.

\subsection{Event Direction (EVT)}

Event studies analyze market returns around dated disclosures~\citep{mackinlay1997eventstudies}.
EVT turns this retrospective analysis into a forecast of the market's reaction over a specified trading window.
EVT supplies an earnings, policy, or macro event, a target asset, a cutoff, and the specified trading session and return definition.
The system predicts \texttt{direction} and \texttt{probability\_up}.
For realized positive-direction label $y_i\in\{0,1\}$ and predicted upward probability $\hat p_i$, the Brier score and EVT score are
\begin{align}
\mathrm{BS} &= \frac{1}{n}\sum_i(\hat p_i-y_i)^2, \\
s_{\mathrm{EVT},c} &= g(1-\mathrm{BS}/0.25).
\end{align}
The score rewards probabilities that assign more weight to the observed direction.
A constant upward probability of 0.5 has $\mathrm{BS}=0.25$ and EVT score zero.
This baseline makes a positive score indicate improvement over an uninformative probability forecast.
Here $g$ clips the skill score to $[0,1]$ for aggregation.
Directional accuracy, macro-F1, and calibration error are auxiliary diagnostics.

\textbf{Example.}\par
\textit{Question:} After CATL's 2025 annual report, predict its open-to-close direction in the specified 2026-03-10 trading session.\par
\textit{Prediction:} The direction is \texttt{up}, with \texttt{probability\_up}=0.7.\par
\textit{Reference:} The session return is positive, so $y=1$.\par
\textit{Scoring:} The Brier loss is $(0.7-1)^2=0.09$, giving an EVT score of $1-0.09/0.25=0.64$ for this example.

\subsection{Fundamentals (FUND)}

FUND extends numerical questions about disclosed financial reports toward forecasting a value not yet published at the cutoff~\citep{chen2021finqa}.
This isolates forecasting company fundamentals from predicting market prices.
The input identifies a stock, historical values and their publication times, a target reporting period, a metric definition, and a unit.
The system returns a \texttt{predicted\_value}, which is compared with the subsequently reported value under the same definition.
For prediction $\hat y_i$ and realized value $y_i$, the FUND score is
\begin{align}
a_i &= \frac{|y_i-\hat y_i|}{|y_i|}, \\
u_{\mathrm{FUND},i} &= \operatorname{clip}(1-a_i,-1,1), \\
s_{\mathrm{FUND},c} &= \frac{1}{n}\sum_i u_{\mathrm{FUND},i}.
\end{align}
Higher scores indicate smaller relative numerical errors.
Relative error makes forecasts of financial metrics comparable across firms with different reporting scales.
MAPE, median absolute percentage error, and the within-10\% rate are auxiliary diagnostics.

\textbf{Example.}\par
\textit{Question:} Given Vanke A's disclosed standalone second-quarter 2025 revenue of CNY 67,328,654,484.06, predict the same metric for the standalone fourth quarter.\par
\textit{Prediction:} The \texttt{predicted\_value} is CNY 67,328,654,484.06.\par
\textit{Reference:} The subsequently disclosed fourth-quarter value is CNY 72,044,353,513.39.\par
\textit{Scoring:} The per-instance FUND utility is approximately $0.9345$.

\subsection{Counterfactual (CF)}

CF draws on synthetic temporal-reasoning tests that check whether a model follows supplied context~\citep{fatemi2024testoftime,weng2026temporalbench}.
It adapts this idea to financial decisions by testing whether fictional event conditions change a predicted excess-return direction.
The input provides a stock, its past 30 trading-day closes, a benchmark, and a fictional announcement.
Favorable and adverse announcements are paired by changing the decisive event term with the price history held fixed.
The system returns a conditional excess-return \texttt{direction} and one-sentence \texttt{reasoning}.
Some instances also request a confidence value and a driver code, while the primary CF score uses the direction.
The evaluator sets the expected direction from the injected event logic.
For predicted direction $\hat d_i$ and reference direction $d_i$, the CF score is
\begin{equation}
s_{\mathrm{CF}}=\mathrm{LA}=\frac{1}{n}\sum_i\mathbf{1}[\hat d_i=d_i].
\end{equation}
Higher logic adherence indicates that responses follow the event specified in the question.
Pair accuracy checks both directions in a favorable--adverse pair, and \texttt{trend\_conflict} adherence checks events whose logic opposes the supplied historical price trend:
Let $\mathcal P$ contain the favorable--adverse pairs and $\mathcal C$ the trend-conflict instances.
\begin{align}
 b_p &= \mathbf{1}[\hat d_{p^+}=d_{p^+}]
        \mathbf{1}[\hat d_{p^-}=d_{p^-}], \\
\mathrm{PairAcc} &= \frac{1}{|\mathcal P|}\sum_{p\in\mathcal P} b_p, \\
\mathrm{LA}_{\mathrm{conflict}} &= \frac{1}{|\mathcal C|}
 \sum_{i\in\mathcal C}\mathbf{1}[\hat d_i=d_i].
\end{align}

\textbf{Example.}\par
\textit{Question:} CATL's supplied 30-day price sequence rises from 160.09 to 190.16; a fictional disclosure-compliance investigation is then injected.\par
\textit{Prediction:} The excess-return direction is \texttt{negative}, with reasoning based on legal and reputational risk.\par
\textit{Reference:} The adverse event's expected logic direction is \texttt{negative}.\par
\textit{Scoring:} This prediction matches the injected logic despite the rising price history, making it a \texttt{trend\_conflict} case.\par

\textbf{Paired example.}\par
\textit{Question:} Given the same Apple price history and S\&P 500 benchmark, predict the five-day excess-return direction after each of two fictional share-repurchase announcements.\par
\textit{Cash-funded announcement:} The company would repurchase up to 5\% of its shares using idle cash, with a price cap 8\% above the current price and no new debt.\par
\textit{Loan-funded announcement:} The repurchase would instead use a two-year pledged loan at 13\% interest, with a price cap 80\% above the current price and a risk to control rights.\par
\textit{Predictions:} Claude Sonnet 4.5 and Gemini 3.1 Flash-Lite both predict \texttt{negative} excess returns after each announcement.\par
\textit{References:} The expected directions are \texttt{positive} for the cash-funded announcement and \texttt{negative} for the loan-funded announcement, based on financing cost and feasibility.\par
\textit{Scoring:} Each model matches one of the two directions, but neither predicts both directions correctly, giving this pair an accuracy of zero.

\subsection{Calculation (CALC)}

CALC tests the financial formula and numerical operations used in research, following the motivation of FinanceReasoning~\citep{tang2025financereasoning}.
Step-level scoring distinguishes mistakes in intermediate calculations from errors in the final answer.
Its input is a fully specified calculation with a topic, difficulty, numerical parameters, and possible distractors or unit traps.
The system returns numerical calculation steps, a final \texttt{answer}, and a \texttt{unit}.
Some calculation templates also request a formula name.
Exact Match is evaluated at every numerical step, including the final answer, against that step's reference value.
Let $\mathcal I_{\mathrm{CALC}}$ contain the calculation problems and $J_i$ be the number of scored numerical steps in problem $i$, including the final answer.
The per-step match indicator, problem score, and CALC score are
\begin{align}
m_{ij} &= \operatorname{match}_{\tau}(\hat z_{ij},z_{ij}), \\
a_i^{\mathrm{CALC}} &= \frac{1}{J_i}\sum_{j=1}^{J_i}m_{ij}, \\
\mathrm{ExactMatch}_{\tau} &= \frac{1}{|\mathcal I_{\mathrm{CALC}}|}
 \sum_{i\in\mathcal I_{\mathrm{CALC}}}a_i^{\mathrm{CALC}}, \\
s_{\mathrm{CALC}} &= \mathrm{ExactMatch}_{\tau}.
\end{align}
The match indicator equals one when the step's numerical error is below the tolerance and its unit is correct, and zero otherwise.
The default tolerance is $\tau=0.002$.
A higher Exact Match score means that more calculation steps are close to their reference values.
For templates that request a formula name, formula accuracy separately records whether it matches its reference.

\textbf{Example.}\par
\textit{Question:} Given next-year free cash flow of CNY 92 million, WACC of 11\%, and perpetual growth of 2\%, calculate terminal enterprise value using the Gordon Growth Model.\par
\textit{Prediction:} The system selects \texttt{gordon\_growth} and predicts a final \texttt{answer} of CNY 1,022.22 million.\par
\textit{Reference:} Both the formula and final answer match their reference values.\par
\textit{Scoring:} The final answer matches its reference as one step; the full Exact Match score also requires the intermediate steps.\par

\textbf{Multi-step example.}\par
\textit{Question:} A company has a 2.4\% risk-free rate, beta of 1.06, and 9\% expected market return.
Its capital structure has 66\% equity and 34\% debt, with a 4.4\% pretax debt cost and a 25\% tax rate.
Given current free cash flow of CNY 620 million, perpetual growth of 2.3\%, net debt of CNY 1.4 billion, and 340 million shares, compute equity value per share.\par
\textit{Prediction:} Claude Sonnet 4.5 reports a 9.396\% cost of equity, 7.32336\% after-tax WACC, CNY 634.26 million in next-year free cash flow, and CNY 11.229 billion in equity value.
Its final estimate is CNY 33.03 per share.\par
\textit{Reference steps:} Cost of equity 9.396\%, after-tax WACC 7.32336\%, next-year free cash flow CNY 634.26 million, and equity value CNY 11.2262103 billion.\par
\textit{Final reference:} CNY 33.02 per share.\par
\textit{Scoring:} The four reference steps and final answer all match within the 0.2\% tolerance.

\subsection{Score Aggregation}

For each forecasting task $t$ and market $c$, let $s_{t,c}$ be the score defined above and $n_{t,c}$ its number of scoreable instances.
The pooled task score weights market scores by their valid instance counts:
\begin{equation}
s_t^{\mathrm{pool}}=
\frac{\sum_{c\in\{\mathrm{CN,US,HK}\}}n_{t,c}s_{t,c}}
{\sum_{c\in\{\mathrm{CN,US,HK}\}}n_{t,c}}.
\end{equation}
For example, market scores of 0.80, 0.70, and 0.60 on 100, 50, and 50 instances give a pooled score of 0.725.
The forecasting Overall Score gives equal weight to the four pooled task scores:
\begin{equation}
S_{\mathrm{forecast}}=
\frac{s_{\mathrm{VAL}}^{\mathrm{pool}}+
s_{\mathrm{RANK}}^{\mathrm{pool}}+
s_{\mathrm{EVT}}^{\mathrm{pool}}+
s_{\mathrm{FUND}}^{\mathrm{pool}}}{4}.
\end{equation}
CF and CALC are reported separately as capability diagnostic tasks.
Overall Score does not average across temporal regimes.

\section{Experimental Details}
\label{app:experiment-details}

This appendix gives the diagnostic metrics and detailed results behind the temporal, market, robustness, and attribution analyses in Section~\ref{sec:experiments}.
The tables report available scores at their stated precision.
A dash (--) denotes a scope without reported T3 results.

\subsection{Forecasting Attribution and Robustness Metrics}
\label{app:metric-formulas}

\paragraph{Leakage and temporal metrics.}
For page-level leakage detection:
\begin{align}
    P &= \frac{\mathrm{TP}}{\mathrm{TP}+\mathrm{FP}}, \\
    R &= \frac{\mathrm{TP}}{\mathrm{TP}+\mathrm{FN}}, \\
    F_1 &= \frac{2PR}{P+R}.
\end{align}
The main diagnostic quantities used in the experimental tables are:
\begin{align}
    \Delta(\mathrm{T1,T2}) &= S_{\mathrm{T1}} - S_{\mathrm{T2}}, \\
    \Delta(\mathrm{T2,T3}) &= S_{\mathrm{T2}} - S_{\mathrm{T3}}.
\end{align}
For robustness, let $\mathcal{P}_p$ and $\mathcal{P}_n$ denote the valid prompt and numerical comparison pairs in a fixed task--market scope.
For a base response $\hat{y}^{0}_i$ and variant $v$, define material-change indicators:
\begin{align}
    C^{p}_{i,v} &= \mathbf{1}\bigl[
        d_p(\hat{y}^{p,v}_i,\hat{y}^{0}_i)>\tau_{p,t}\bigr], \\
    C^{n}_{i,v} &= \mathbf{1}\bigl[
        d_{n,v}(\hat{y}^{n,v}_i,\hat{y}^{0}_i)>\tau_{n,t}\bigr].
\end{align}
The conditional change rates are:
\begin{align}
    \mathrm{PCR} &= \frac{1}{|\mathcal{P}_p|}
        \sum_{(i,v)\in\mathcal{P}_p}C^p_{i,v}, \\
    \mathrm{NCR} &= \frac{1}{|\mathcal{P}_n|}
        \sum_{(i,v)\in\mathcal{P}_n}C^n_{i,v}.
\end{align}
Here PCR is Prompt Change Rate and NCR is Perturb Change Rate.
The denominator is the number of valid comparison pairs, not automatically the number of base instances times the number of proposed variants.
A scope with no valid pairs has an unavailable rate, not a zero rate.

Let $s^0_{i,v}$ and $s^{p,v}_i$ be the original and paraphrased per-example scores evaluated on the same valid pair, with analogous scores for numerical variants.
The mean score changes and combined Score Delta are:
\begin{align}
    \Delta_p S &= \frac{1}{|\mathcal{P}_p|}
        \sum_{(i,v)\in\mathcal{P}_p}(s^{p,v}_i-s^0_{i,v}), \\
    \Delta_n S &= \frac{1}{|\mathcal{P}_n|}
        \sum_{(i,v)\in\mathcal{P}_n}(s^{n,v}_i-s^0_{i,v}), \\
    \mathrm{ScoreDelta} &= (\Delta_p S+\Delta_n S)/2.
\end{align}
The quantities $\Delta_p S$ and $\Delta_n S$ are intermediate components of the reported Score Delta.

The task-specific comparison functions use direction changes for EVT/CF, rank distance for RANK, and numerical tolerances for VAL/FUND/CALC.
Numerical comparison accounts for unit conversion and permissible response to genuine input changes; formula validity in CALC is checked against the appropriate reference, not literal expression identity.
For CALC input perturbations, reference answers are recomputed, so a correct numerical response is not penalized merely for differing from the original answer.
\paragraph{T2--T3 proximity.}
Delta(T2,T3) is positive when the backtest score exceeds the live score and negative when it falls below the live score.
Its absolute value describes the observed score gap within the reported model--task--market scope.
A small gap indicates closer backtest-to-live agreement for that model, task, and market.

\subsection{Temporal Results}

Historical scores exceed post-cutoff scores in 25 of the 27 RANK, EVT, and FUND model--task--market comparisons, pointing to an advantage from historical knowledge.
Every RANK and FUND gap is positive; the two remaining cases are HK EVT.
The CN FUND gaps range from 0.3503 to 0.4469, making historical advantage especially pronounced for that task and market.
Table~\ref{tab:app-main-full-matrix} gives the T1 and T2 scores behind Table~\ref{tab:temporal-summary}, together with VAL scores and the available US T3 references.
The focused comparison uses RANK, EVT, and FUND because VAL is more sensitive to short-horizon price fluctuations.
Table~\ref{tab:app-filter-transfer} then compares the same US T3 reference within each model--task pair against three T2 retrieval settings over matched one-week windows.

\begin{table*}[!htbp]
\centering
\scriptsize
\setlength{\tabcolsep}{4pt}
\begin{tabular*}{\textwidth}{@{\extracolsep{\fill}}lllrrrr@{}}
\toprule
\textbf{Model} & \textbf{Task} & \textbf{Market} & \textbf{T1 Score} & \textbf{T2 Score} & \shortstack{\textbf{T3 Score}\\\textbf{(reference)}} & \textbf{Delta(T1,T2)} \\
\midrule
\multirow{12}{*}{GPT-4.1-mini} & VAL & CN & 0.8153 & 0.9028 & -- & -0.0875 \\
 & VAL & US & 0.7827 & 0.8502 & 0.8417 & -0.0675 \\
 & VAL & HK & 0.6552 & 0.7107 & -- & -0.0555 \\
 & RANK & CN & 0.5240 & 0.4676 & -- & 0.0564 \\
 & RANK & US & 0.5172 & 0.4538 & 0.4501 & 0.0635 \\
 & RANK & HK & 0.5144 & 0.4798 & -- & 0.0346 \\
 & EVT & CN & 0.7368 & 0.6819 & -- & 0.0549 \\
 & EVT & US & 0.6446 & 0.5847 & 0.5805 & 0.0599 \\
 & EVT & HK & 0.6903 & 0.7021 & -- & -0.0117 \\
 & FUND & CN & 0.9114 & 0.4644 & -- & 0.4469 \\
 & FUND & US & 0.8380 & 0.7335 & 0.7194 & 0.1045 \\
 & FUND & HK & 0.9259 & 0.7657 & -- & 0.1602 \\
\midrule
\multirow{12}{*}{Claude Sonnet 4.5} & VAL & CN & 0.7126 & 0.8847 & -- & -0.1721 \\
 & VAL & US & 0.7975 & 0.8456 & 0.8296 & -0.0481 \\
 & VAL & HK & 0.5725 & 0.7673 & -- & -0.1948 \\
 & RANK & CN & 0.5669 & 0.4534 & -- & 0.1135 \\
 & RANK & US & 0.5703 & 0.4938 & 0.4249 & 0.0765 \\
 & RANK & HK & 0.6194 & 0.5201 & -- & 0.0992 \\
 & EVT & CN & 0.8461 & 0.7233 & -- & 0.1227 \\
 & EVT & US & 0.7115 & 0.6006 & 0.5926 & 0.1108 \\
 & EVT & HK & 0.7578 & 0.7096 & -- & 0.0482 \\
 & FUND & CN & 0.9997 & 0.5655 & -- & 0.4341 \\
 & FUND & US & 0.9689 & 0.8829 & 0.8563 & 0.0859 \\
 & FUND & HK & 0.9832 & 0.7650 & -- & 0.2182 \\
\midrule
\multirow{12}{*}{Gemini 3.1 Flash-Lite} & VAL & CN & 0.8384 & 0.8781 & -- & -0.0396 \\
 & VAL & US & 0.9146 & 0.8885 & 0.8678 & 0.0261 \\
 & VAL & HK & 0.6950 & 0.6921 & -- & 0.0029 \\
 & RANK & CN & 0.5859 & 0.4953 & -- & 0.0906 \\
 & RANK & US & 0.6845 & 0.5136 & 0.4272 & 0.1709 \\
 & RANK & HK & 0.6424 & 0.4868 & -- & 0.1555 \\
 & EVT & CN & 0.8228 & 0.7215 & -- & 0.1013 \\
 & EVT & US & 0.7152 & 0.6893 & 0.6790 & 0.0259 \\
 & EVT & HK & 0.7496 & 0.7575 & -- & -0.0079 \\
 & FUND & CN & 0.9966 & 0.6463 & -- & 0.3503 \\
 & FUND & US & 0.9533 & 0.8636 & 0.8292 & 0.0897 \\
 & FUND & HK & 0.9769 & 0.8246 & -- & 0.1523 \\
\bottomrule
\end{tabular*}
\caption{Available FinanceHarness temporal scores.
T3 reference scores are available for the 12 US model--task combinations.
Dashes (--) mark CN/HK scopes without reported T3 results, not observed forecasting failures.
Small differences between listed deltas and subtraction of displayed scores can reflect rounding.}
\label{tab:app-main-full-matrix}
\end{table*}

\begin{table*}[!htbp]
\centering
\footnotesize
\setlength{\tabcolsep}{6pt}
\renewcommand{\arraystretch}{1.08}

\begin{tabular*}{\textwidth}{
@{\extracolsep{\fill}}
lllrrr
@{}
}
\toprule
\textbf{Model}
& \textbf{Scope}
& \textbf{Search Setting}
& \textbf{T2 Score}
& \textbf{T3 Score}
& \textbf{Delta(T2,T3)} \\
\midrule

\multirow{9}{*}{GPT-4.1-mini}
& \multirow{3}{*}{RANK + US}
& search only
& 0.4633
& \multirow{3}{*}{0.4501}
& 0.0132 \\

&
& search + PIT
& 0.4594
&
& 0.0093 \\

&
& search + PIT + LLM Filter
& 0.4538
&
& 0.0037 \\

\cmidrule(lr){2-6}

& \multirow{3}{*}{EVT + US} & search only & 0.6464 & \multirow{3}{*}{0.5805} & 0.0659 \\
& & search + PIT & 0.6489 & & 0.0684 \\
& & search + PIT + LLM Filter & 0.5847 & & 0.0043 \\
\cmidrule(lr){2-6}

&
\multirow{3}{*}{FUND + US}
& search only
& 0.8530
& \multirow{3}{*}{0.7194}
& 0.1337 \\

&
& search + PIT
& 0.7522
&
& 0.0328 \\

&
& search + PIT + LLM Filter
& 0.7335
&
& 0.0141 \\

\midrule

\multirow{9}{*}{Gemini 3.1 Flash-Lite}
& \multirow{3}{*}{RANK + US}
& search only
& 0.5557
& \multirow{3}{*}{0.4272}
& 0.1286 \\

&
& search + PIT
& 0.5019
&
& 0.0748 \\

&
& search + PIT + LLM Filter
& 0.5136
&
& 0.0864 \\

\cmidrule(lr){2-6}

& \multirow{3}{*}{EVT + US} & search only & 0.7867 & \multirow{3}{*}{0.6790} & 0.1077 \\
& & search + PIT & 0.6951 & & 0.0161 \\
& & search + PIT + LLM Filter & 0.6893 & & 0.0103 \\
\cmidrule(lr){2-6}

&
\multirow{3}{*}{FUND + US}
& search only
& 0.9756
& \multirow{3}{*}{0.8292}
& 0.1465 \\

&
& search + PIT
& 0.8764
&
& 0.0473 \\

&
& search + PIT + LLM Filter
& 0.8636
&
& 0.0344 \\

\midrule

\multirow{9}{*}{Claude Sonnet 4.5}
& \multirow{3}{*}{RANK + US}
& search only
& 0.5495
& \multirow{3}{*}{0.4249}
& 0.1246 \\

&
& search + PIT
& 0.4786
&
& 0.0537 \\

&
& search + PIT + LLM Filter
& 0.4938
&
& 0.0689 \\

\cmidrule(lr){2-6}

& \multirow{3}{*}{EVT + US} & search only & 0.8232 & \multirow{3}{*}{0.5926} & 0.2306 \\
& & search + PIT & 0.7808 & & 0.1882 \\
& & search + PIT + LLM Filter & 0.6006 & & 0.0080 \\
\cmidrule(lr){2-6}

&
\multirow{3}{*}{FUND + US}
& search only
& 0.9882
& \multirow{3}{*}{0.8563}
& 0.1320 \\

&
& search + PIT
& 0.9221
&
& 0.0659 \\

&
& search + PIT + LLM Filter
& 0.8829
&
& 0.0267 \\

\bottomrule
\end{tabular*}
\caption{US RANK, EVT, and FUND T2/T3 scores under three search settings in FinanceHarness with matched one-week evaluation windows.
Only T2 settings vary against the shared T3 reference within each model--task scope.
Scores and gaps are shown to four decimal places.
Small differences between a reported gap and subtraction of displayed scores reflect rounding.}
\label{tab:app-filter-transfer}
\end{table*}

The main-text gap reduction is computed from the unweighted mean absolute T2--T3 gap over the nine model--task pairs in Table~\ref{tab:app-filter-transfer}.
The mean falls from 0.1203 with search alone to 0.0618 with PIT and 0.0285 with PIT plus filtering, a reduction of approximately 76\% from the search-only setting.
Filtering gives the closest live agreement for every EVT and FUND comparison, and PIT alone yields the smallest RANK gap for Claude Sonnet 4.5 and Gemini 3.1 Flash-Lite.
The controlled T2 scores therefore provide timely feedback that closely tracks live EVT and FUND performance, with more task-dependent agreement for RANK.

\subsection{Market-Level Analysis}
\label{app:market-level-analysis}

Forecasting strengths change across markets and model--harness configurations.
Table~\ref{tab:market-summary} shows that all six models have their highest three-harness mean in the US, yet DeepSeek V4 Flash leads in CN and Gemini 3.1 Flash-Lite leads in US and HK.
Table~\ref{tab:app-market-details} reports the 54 model--harness market scores behind those means, with each row comparing the same system across CN, US, and HK.
The harness-level pattern adds another distinction: GPT-4.1-mini scores highest in HK with FinanceHarness and Open Deep Research, and highest in US with GPT-Researcher.
Market-specific results thus reveal both changes in the leading model and shifts within the same model across research workflows.

\begin{table}[!htbp]
\centering
\footnotesize
\renewcommand{\arraystretch}{1.15}
\setlength{\tabcolsep}{3pt}
\begin{tabularx}{\columnwidth}{Xrrr}
\toprule
\rowcolor{tableHeader}
\textbf{Model} & \textbf{CN} & \textbf{US} & \textbf{HK} \\
\midrule
GPT-4.1-mini & 0.6173 & 0.6536 & 0.6341 \\
Gemini 3.1 Flash-Lite & 0.6772 & 0.7480 & 0.6698 \\
Claude Sonnet 4.5 & 0.6380 & 0.7262 & 0.6679 \\
DeepSeek V4 Flash & 0.6825 & 0.7235 & 0.6613 \\
Phi-4 & 0.6618 & 0.6803 & 0.6246 \\
Gemma-3 & 0.5628 & 0.6366 & 0.5659 \\
\bottomrule
\end{tabularx}
\vspace{-0.2cm}
\caption{Model-level T2 market scores averaged equally across the three research harnesses within each market.}
\label{tab:market-summary}
\vspace{-0.2cm}
\end{table}

\begin{table}[!htbp]
\centering
\footnotesize
\setlength{\tabcolsep}{2pt}
\renewcommand{\arraystretch}{1.08}
\begin{tabularx}{\columnwidth}{Xlrrr}
\toprule
\rowcolor{tableHeader}
\textbf{Model} & \textbf{Harness} & \textbf{CN} & \textbf{US} & \textbf{HK} \\
\midrule
\multirow{3}{*}{GPT-4.1-mini} & FH & 0.6292 & 0.6556 & \bestscore{0.6646} \\
 & ODR & 0.5701 & 0.6372 & \bestscore{0.6480} \\
 & GPT-R & 0.6525 & \bestscore{0.6679} & 0.5898 \\
\midrule
\multirow{3}{*}{\shortstack[l]{Gemini 3.1\\Flash-Lite}} & FH & 0.6853 & \bestscore{0.7387} & 0.6903 \\
 & ODR & 0.6527 & \bestscore{0.7437} & 0.6736 \\
 & GPT-R & 0.6937 & \bestscore{0.7616} & 0.6454 \\
\midrule
\multirow{3}{*}{\shortstack[l]{Claude Sonnet\\4.5}} & FH & 0.6567 & \bestscore{0.7057} & 0.6905 \\
 & ODR & 0.6493 & \bestscore{0.7612} & 0.6378 \\
 & GPT-R & 0.6080 & \bestscore{0.7118} & 0.6754 \\
\midrule
\multirow{3}{*}{\shortstack[l]{DeepSeek V4\\Flash}} & FH & 0.6831 & \bestscore{0.7142} & 0.6437 \\
 & ODR & 0.6553 & \bestscore{0.7365} & 0.6644 \\
 & GPT-R & 0.7091 & \bestscore{0.7199} & 0.6758 \\
\midrule
\multirow{3}{*}{Phi-4} & FH & \bestscore{0.7249} & 0.6390 & 0.6355 \\
 & ODR & 0.6642 & \bestscore{0.7064} & 0.6619 \\
 & GPT-R & 0.5964 & \bestscore{0.6955} & 0.5765 \\
\midrule
\multirow{3}{*}{Gemma-3} & FH & 0.5723 & \bestscore{0.6099} & 0.5761 \\
 & ODR & 0.5789 & \bestscore{0.6501} & 0.6110 \\
 & GPT-R & 0.5371 & \bestscore{0.6497} & 0.5107 \\
\bottomrule
\end{tabularx}
\caption{Complete market-level T2 results underlying Table~\ref{tab:market-summary}.
FH denotes FinanceHarness, ODR Open Deep Research, and GPT-R GPT-Researcher.
Bold type marks the highest market score within each model--harness row.}
\label{tab:app-market-details}
\end{table}

\subsection{Robustness Details}

Robustness results distinguish changes in individual predictions from changes in aggregate scores.
Table~\ref{tab:robustness-summary} averages EVT and FUND results across CN, US, and HK for GPT-4.1-mini and Claude Sonnet 4.5 in two harnesses.
For both models, FinanceHarness has lower Prompt and Perturb Change Rates than GPT-Researcher.
Across the 24 task--market rows in Table~\ref{tab:app-robustness-observed}, prompt rewrites change outputs more often than numerical perturbations.
The difference is visible for GPT-4.1-mini on HK FUND with GPT-Researcher: its Prompt Change Rate is 0.35 and Perturb Change Rate is 0.10, with a Score Delta of -0.0022592.
All four aggregate Score Deltas have magnitude below 0.0026, showing that these response changes have a much smaller effect on mean task scores.

\begin{table}[!htbp]
\centering
\footnotesize
\setlength{\tabcolsep}{3pt}
\renewcommand{\arraystretch}{1.08}
\begin{tabularx}{\columnwidth}{Xlrrr}
\toprule
\rowcolor{tableHeader}
\textbf{Model} & \textbf{Harness} & \shortstack{\textbf{Prompt}\\\textbf{CR}} & \shortstack{\textbf{Perturb}\\\textbf{CR}} & \shortstack{\textbf{Score}\\\textbf{Delta}} \\
\midrule
\multirow{2}{*}{GPT-4.1-mini} & FH & 0.2150 & 0.0867 & -0.002004 \\
 & GPT-R & 0.2367 & 0.0917 & -0.002512 \\
\midrule
\multirow{2}{*}{\shortstack[l]{Claude Sonnet\\4.5}} & FH & 0.1683 & 0.0700 & -0.001473 \\
 & GPT-R & 0.1883 & 0.0783 & -0.002017 \\
\bottomrule
\end{tabularx}
\vspace{-0.2cm}
\caption{Mean T2 robustness across EVT/FUND and CN/US/HK.
CR denotes Change Rate, FH denotes FinanceHarness, and GPT-R denotes GPT-Researcher.}
\label{tab:robustness-summary}
\vspace{-0.2cm}
\end{table}

\begin{table*}[!htbp]
\centering
\scriptsize
\setlength{\tabcolsep}{3pt}
\begin{tabular*}{\textwidth}{@{\extracolsep{\fill}}llllrrr@{}}
\toprule
\textbf{Model} & \textbf{Harness} & \textbf{Task} & \textbf{Market} & \shortstack{\textbf{Prompt}\\\textbf{Change Rate}} & \shortstack{\textbf{Perturb}\\\textbf{Change Rate}} & \textbf{Score Delta} \\
\midrule
\multirow{12}{*}{GPT-4.1-mini} & \multirow{6}{*}{FinanceHarness} & EVT & CN & 0.17 & 0.12 & -0.00147222 \\
 & & EVT & US & 0.18 & 0.08 & -0.00384787 \\
 & & EVT & HK & 0.23 & 0.09 & 0.0028771 \\
 & & FUND & CN & 0.23 & 0.07 & -0.0034291 \\
 & & FUND & US & 0.27 & 0.07 & -0.0010748 \\
 & & FUND & HK & 0.21 & 0.09 & -0.0050801 \\
\cmidrule(lr){2-7}
 & \multirow{6}{*}{GPT-Researcher} & EVT & CN & 0.22 & 0.07 & -0.00427368 \\
 & & EVT & US & 0.14 & 0.08 & 0.0003339 \\
 & & EVT & HK & 0.26 & 0.08 & -0.00141261 \\
 & & FUND & CN & 0.2 & 0.13 & -0.0073845 \\
 & & FUND & US & 0.25 & 0.09 & -0.0000735 \\
 & & FUND & HK & 0.35 & 0.1 & -0.0022592 \\
\midrule
\multirow{12}{*}{Claude Sonnet 4.5} & \multirow{6}{*}{FinanceHarness} & EVT & CN & 0.12 & 0.05 & -0.00057057 \\
 & & EVT & US & 0.1 & 0.03 & 0.0002345 \\
 & & EVT & HK & 0.2 & 0.07 & -0.00250851 \\
 & & FUND & CN & 0.17 & 0.09 & -0.0041914 \\
 & & FUND & US & 0.25 & 0.09 & -0.0009102 \\
 & & FUND & HK & 0.17 & 0.09 & -0.0008937 \\
\cmidrule(lr){2-7}
 & \multirow{6}{*}{GPT-Researcher} & EVT & CN & 0.13 & 0.04 & -0.00142874 \\
 & & EVT & US & 0.13 & 0.07 & -0.00075541 \\
 & & EVT & HK & 0.16 & 0.07 & 0.0008813 \\
 & & FUND & CN & 0.21 & 0.1 & -0.0015307 \\
 & & FUND & US & 0.28 & 0.12 & -0.0053947 \\
 & & FUND & HK & 0.22 & 0.07 & -0.0038728 \\
\bottomrule
\end{tabular*}
\caption{Consolidated T2 robustness results across CN/US/HK.
Score Delta is the mean of prompt and numerical score changes.}
\label{tab:app-robustness-observed}
\end{table*}

\subsection{Capability Diagnostic Results}

CF tests supplied event logic and CALC tests stepwise calculation; Table~\ref{tab:main-forecasting-t2} reports both scores for 22 configurations.
Their scoring rules and examples appear in Appendix~\ref{app:task-templates}.
Within FinanceHarness, CF ranges from 0.9000 to 1.0000 across the three models in the temporal analysis.
Across all configurations, CALC spans 0.2143 to 1.0000, showing greater variation in stepwise numerical reliability.
Claude Sonnet 4.5 scores 1.0000 on both capability tasks with Claude Code, yet its forecasting Overall is 0.6292 there and 0.6797 with FinanceHarness.
This contrast separates capability scores from future-outcome prediction.

\newpage
\subsection{Prediction Attribution}
\label{app:prediction-attribution}

Attribution combines historical advantage, retrieval-policy sensitivity, live performance, and input robustness for the same forecasting tasks.
For the four US EVT/FUND comparisons, Table~\ref{tab:e4-attribution-results} follows T2 scores from search alone through PIT and filtering to the T3 reference.
Table~\ref{tab:attribution-summary} adds the T1--T2 and filtered T2--T3 gaps and the combined Input Score Delta.
The broader RANK/EVT/FUND temporal evidence appears in Tables~\ref{tab:app-main-full-matrix} and~\ref{tab:app-filter-transfer}.

\begin{table*}[!htbp]
\centering
\scriptsize
\setlength{\tabcolsep}{3pt}
\renewcommand{\arraystretch}{1.15}
\begin{tabularx}{\textwidth}{lc*{6}{>{\centering\arraybackslash}X}}
\toprule
\rowcolor{tableHeader}
\textbf{Model} & \textbf{Task} & \textbf{Search} & \textbf{+ PIT} & \shortstack{\textbf{+ PIT}\\\textbf{+ Filter}} & \textbf{T3} & \shortstack{\textbf{Delta}\\\textbf{(T1,T2)}} & \shortstack{\textbf{Delta}\\\textbf{(T2,T3)}} \\
\midrule
\multirow{2}{*}{GPT-4.1-mini} & EVT & 0.6464 & 0.6489 & 0.5847 & 0.5805 & +0.0599 & +0.0043 \\
 & FUND & 0.8530 & 0.7522 & 0.7335 & 0.7194 & +0.1045 & +0.0141 \\
\midrule
\multirow{2}{*}{\shortstack[l]{Claude Sonnet\\4.5}} & EVT & 0.8232 & 0.7808 & 0.6006 & 0.5926 & +0.1108 & +0.0080 \\
 & FUND & 0.9882 & 0.9221 & 0.8829 & 0.8563 & +0.0859 & +0.0267 \\
\bottomrule
\end{tabularx}
\caption{Task-specific scores underlying prediction attribution in FinanceHarness on US EVT and FUND.
Search, + PIT, and + PIT + Filter report T2 scores.
Delta(T2,T3) is the filtered T2 score minus the T3 reference score.
Scores and gaps are displayed to four decimal places.
Small differences between a reported gap and subtraction of displayed scores reflect rounding.}
\label{tab:e4-attribution-results}
\end{table*}

\paragraph{GPT-4.1-mini.}
EVT and FUND have positive T1--T2 gaps of 0.0599 and 0.1045, indicating stronger performance on pretraining-era history.
Retrieval control moves their T2 scores from 0.6464 and 0.8530 under search alone to 0.5847 and 0.7335 after filtering.
The filtered scores approach their live T3 scores of 0.5805 and 0.7194, leaving gaps of 0.0043 and 0.0141.
Input Score Deltas of -0.003848 and -0.001075 are small beside the retrieval-policy shifts.
For this model, information access is the dominant observed influence on retrospective EVT/FUND scores, and controlled T2 closely tracks live performance.

\paragraph{Claude Sonnet 4.5.}
The positive T1--T2 gaps of 0.1108 for EVT and 0.0859 for FUND show a historical advantage on both tasks.
EVT exhibits the largest retrieval-policy shift among the four comparisons: its T2 score falls from 0.8232 under search alone to 0.6006 after filtering.
That filtered score closely matches its T3 score of 0.5926, and FUND shows a similar pattern with filtered T2 at 0.8829 and T3 at 0.8563.
The corresponding Input Score Deltas, +0.000235 and -0.000910, are much smaller than the retrieval-policy shifts.
Claude retains substantial live forecasting performance, with retrieval access contributing most to the observed change in its historical EVT/FUND scores.